\documentclass{bmvc2k}

\usepackage{color, colortbl}
\usepackage{hyperref}
\definecolor{Gray}{gray}{0.9}
\makeatletter
\newcommand{\thickhline}{%
    \noalign {\ifnum 0=`}\fi \hrule height 1pt
    \futurelet \reserved@a \@xhline
}

\makeatletter
\renewcommand{\paragraph}{%
  \@startsection{paragraph}{4}%
  {\z@}{0.5 ex \@plus 1ex \@minus .2ex}{-1em}%
  {\normalfont\normalsize\bfseries}%
}
\makeatother

\title{Recognizing and Curating Photo Albums via Event-Specific Image Importance}

\addauthor{Yufei Wang}{yuw176@ucsd.edu}{1}
\addauthor{Zhe Lin}{zlin@adobe.com}{2}
\addauthor{Xiaohui Shen}{xshen@adobe.com}{2}
\addauthor{Radom\'{i}r M\v{e}ch}{rmech@adobe.com}{2}
\addauthor{Gavin Miller}{gmiller@adobe.com}{2}
\addauthor{Garrison W. Cottrell}{gary@ucsd.edu}{1}

\addinstitution{
 University of California, San Diego\\
 9500 Gilman Drive, \\
 San Diego, USA
}
\addinstitution{
 Adobe Research \\
 345 Park Avenue, \\
 San Jose, USA
}

\runninghead{W. Yufei \etal}{Recognizing and Curating Photo Albums}

\def\etal{\emph{et al}\bmvaOneDot}

\begin{document}

\maketitle

\begin{abstract}
Automatic organization of personal photos is a problem with many real world applications, and can be divided into two main tasks: recognizing the event type of the photo collection, and selecting interesting images from the collection. In this paper, we attempt to simultaneously solve both tasks: album-wise event recognition and image-wise importance prediction. We collected an album dataset with both event type labels and image importance labels, refined from an existing CUFED dataset. We propose a hybrid system consisting of three parts: A siamese network-based event-specific image importance prediction, a Convolutional Neural Network (CNN) that recognizes the event type, and a Long Short-Term Memory (LSTM)-based sequence level event recognizer. We propose an iterative updating procedure for event type and image importance score prediction. We experimentally verified that image importance score prediction and event type recognition can each help the performance of the other. 
\end{abstract}

\section{Introduction}
\vspace{-5pt}
With the advent of cheap cameras in nearly all of our devices,  automated uploading to the cloud, and practically unlimited storage,
it has become painless to take photos frequently in daily life, resulting in an explosion of personal photo collections. However, the oversized image collections make it difficult to organize the photos, and thus automatic organization algorithms are highly desirable. The organization of personal photo collections can be decomposed into two stages: recognizing the event type of a photo collection, and suggesting the most interesting/important images in the photo collection to represent the album. 
The two stages can assist users in keeping the photo collections organized and free of irrelevant images, and can be further used to pick photos for an album cover or to make a photo collage.

Both event recognition and image importance prediction have been studied independently in previous literature. Studies of event recognition fall into three types. The most popular approach uses videos as input \cite{video16, 2015trecvidover, TangCVPR12, xu2015discriminative, Ng_2015_CVPR, NagelBMVC2015, devnet}, and spatiotemporal features are commonly used. Further, event recounting which aims to find the event-specific spatial/temporal discriminative parts of a video is also studied \cite{devnet}. This is relevant to event-specific image importance, but the image importance of an image is not decided by how discriminative it is.
At the other end of the spectrum, event recognition for single images has also been studied \cite{what_where, Park_2015_CVPR_Workshops, cSalvadora, WangW0G16}. 
There is no temporal information or relevant frame importance to consider, and both object and scene level features have been used \cite{what_where}. 

Album-wise event recognition lies between single-image-based and video-based event recognition, and is most related to our work. Images in an album can be thought of as very sparse samples from an event video, and consecutive images from the photo album are no longer continuous. 
A common approach is to aggregate evidence from single images to classify the album type \cite{Mattivi11, pattern,  icpr2016, pec22}. For example,  Wu \etal \cite{pec22} fine-tune Alexnet to extract features from single image, and then aggregate the features from each image and train a multi-layer network to recognize the event type of the album. The above work treats albums as unordered collection of images. On the other hand, Bossard \etal \cite{HMM} exploit the sequential nature of personal albums, using an HMM-based sub-event approach (Stopwatch HMM) for event recognition. They use temporal sequence of the images, and model an album with successive latent sub-events to boost the recognition performance, and show that the temporally-sensitive HMM outperforms simply aggregating the predictions from all the images in an album. This indicates that the sequential information in an album is useful for  album-wise event recognition.

Image importance is a complex image property which is related to various factors, such as aesthetics \cite{aesthe_14}, interestingness \cite{interesting} and image memorability \cite{Isola2011}. Wang \etal \cite{CVPR} show that image importance is modulated by the context it is in, i.e., image importance is event-specific. For example, a photo of a beautiful work of architecture is important in an album of an urban trip, yet not so important in a wedding event. 
They showed that a siamese-network-based model can reasonably predict this highly subjective image property. However, their work assumed that the event type of the album is already known. This is undesirable if we want to build an end-to-end photo organization algorithm. In this work, we train a system simultaneously for event recognition and image curation, so that user input of the event type is not required.

Event recognition and image importance prediction are inherently related to each other: 1) importance is event-specific, so we need to know the event type to better predict importance; 2) albums often contain ``outlier'' photos that aren't directly related to the event.
If we can reduce effects from the outliers by discovering the important/key images in an album, we can better recognize the event.  Therefore, we ask the question: can we simultaneously recognize the event type of an album, and discover important images in it? And more importantly, can we improve the performance of each task by forming a joint solution?

In answering this question, this paper makes the following contributions: 1) We develop a joint event recognition and image importance prediction algorithm.
We use a CNN for image level event recognition, and a Siamese Network for event-specific image importance prediction. Then an iterative update scheme is used during the test stage, and we find that event recognition and image importance prediction can indeed improve each other's performance; 2) We further boost the performance of event recognition with an LSTM network that leverages sequential information in labeling the album; 3) We also refine the CUFED dataset by collecting more human annotations for the event types, allowing raters to apply multiple labels to the events. This improves the reliability of the ground-truth, accounting for the ambiguity between event types. 

\section{The ML-CUFED Dataset}
\vspace{-5pt}

    In order to train and evaluate the joint curation-recognition model, we use the Curation of Flickr Events Dataset (CUFED) \cite{CVPR}, and refine it by collecting additional human opinions on the event types in the dataset. We call the new dataset MultiLabel-CUFED (ML-CUFED). In this section, we describe the dataset, and provide a consistency analysis of the labels collected from Amazon Mechanical Turk (AMT). The dataset is publicly available\footnote{
\url{http://acsweb.ucsd.edu/~yuw176/event-curation.html}}. More details of ML-CUFED are in the supplementary material.
\vspace{-5pt}
\subsection{The CUFED dataset}
\vspace{-5pt}

The CUFED dataset \cite{CVPR} is an image curation dataset extracted from the Yahoo Flickr Creative Commons 100M dataset.  It contains 1883 albums over 23 common event types, with 50 to 200 albums for each event type. The event type of each album was decided by 3 AMT workers' annotations. Meanwhile, within each album, the event-specific importance of each image is obtained by averaging 5 AMT workers' votes when the event type is given to them. 

One problem with CUFED is that the event type of an album is decided by only 3 workers, who were constrained to give a single label to each album. However, some of the event types in that dataset are related (e.g., architecture and urban trip). For an album with ambiguous or multiple event types, such a constraint is overly restrictive. 
Therefore, collecting the event types and their proportion in one album from more peoples' views is necessary. This results in a multi-label event recognition dataset with richer information. In the supplementary material, we show examples of event labels from CUFED and our refined labels.


\subsection{Data collection and Analysis}
\vspace{-3pt}
In addition to the 3 votes the dataset already includes, we collected 9 more workers' opinions for each album, and allowed them to select up to 3 event types. There were  299 distinct workers who participated in the task.

Quality control was performed for each AMT worker in order to collect high quality annotations. Before the real task, only workers who passed a test that was very similar to the actual task were allowed to proceed. 
During the tasks, 
the results workers turned in were compared with other workers' submissions, and submissions that highly diverged from others were further manually inspected. If the divergence was unreasonable, the submission was rejected. After all the annotations from workers were collected, we further cleaned the annotations by eliminating the labels with only one vote. To get the final ground-truth event types, we converted the votes to a probability distribution over event types for an album. 

To check the validity of the dataset we collected, we analyzed the annotations in several ways. Each album has between 9 and 27 votes (because we allow for multiple choices from one worker). 76\% of the albums received votes for two or fewer event types. 
95\% of the albums received votes for three or fewer event types.  To check the consistency among workers, we randomly split the 299 workers into two halves, and for each album we checked whether the annotations from one half were consistent with the other half. 
We repeated the random split 100 times, and on average, for 89.6\% of the albums, the event type receiving the most votes were the same for both groups. This suggests that despite the ambiguity of some album types, the opinions of different AMT workers are consistent.

\section{Joint Event Recognition and Image Curation}
\begin{figure}
\centering
\includegraphics[width=1\textwidth]{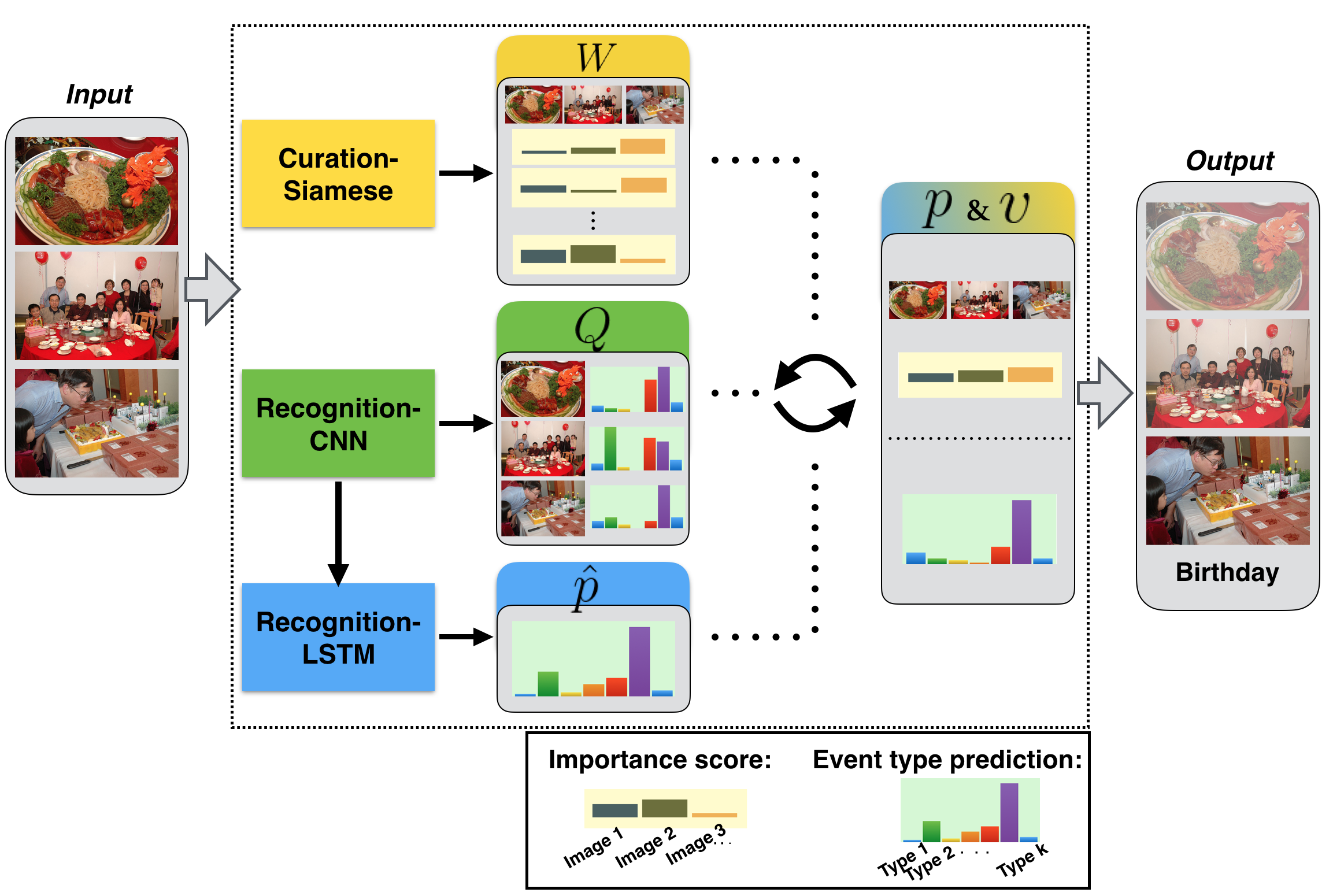}
\vspace{-0.2in}
\caption{The joint album recognition-curation system. \{$W, Q, \hat{p}, p, v$\} are described in Section~\ref{iterative-section}. $W$, $Q$, and $\hat{p}$ are computed once and then used to iteratively update $p$ and $v$.
}
\label{figure1}
\vspace{-1pt}
\end{figure}

\vspace{-5pt}
\label{approach}
In this section, we describe our approach to jointly attain image importance prediction and album event recognition. It is intuitive that important images contribute more to the identity of an event, and should be emphasized when deciding the event type of the album from the images. On the other hand, the identity of the event is needed for accurate individual image importance prediction, as shown in \cite{CVPR}. Moreover, it has been shown that sequential information in an album is useful for event prediction \cite{HMM}. Therefore, we build a joint system that can simultaneously predict the event type and image importance for an album. The system is shown in Figure~\ref{figure1}. We  elaborate on the different parts of the system in this section.

\vspace{-2pt}
\subsection{Event curation network}
\vspace{-2pt}
\label{curation_section}

\begin{figure}[ht]
\centering
\includegraphics[width=1\textwidth]{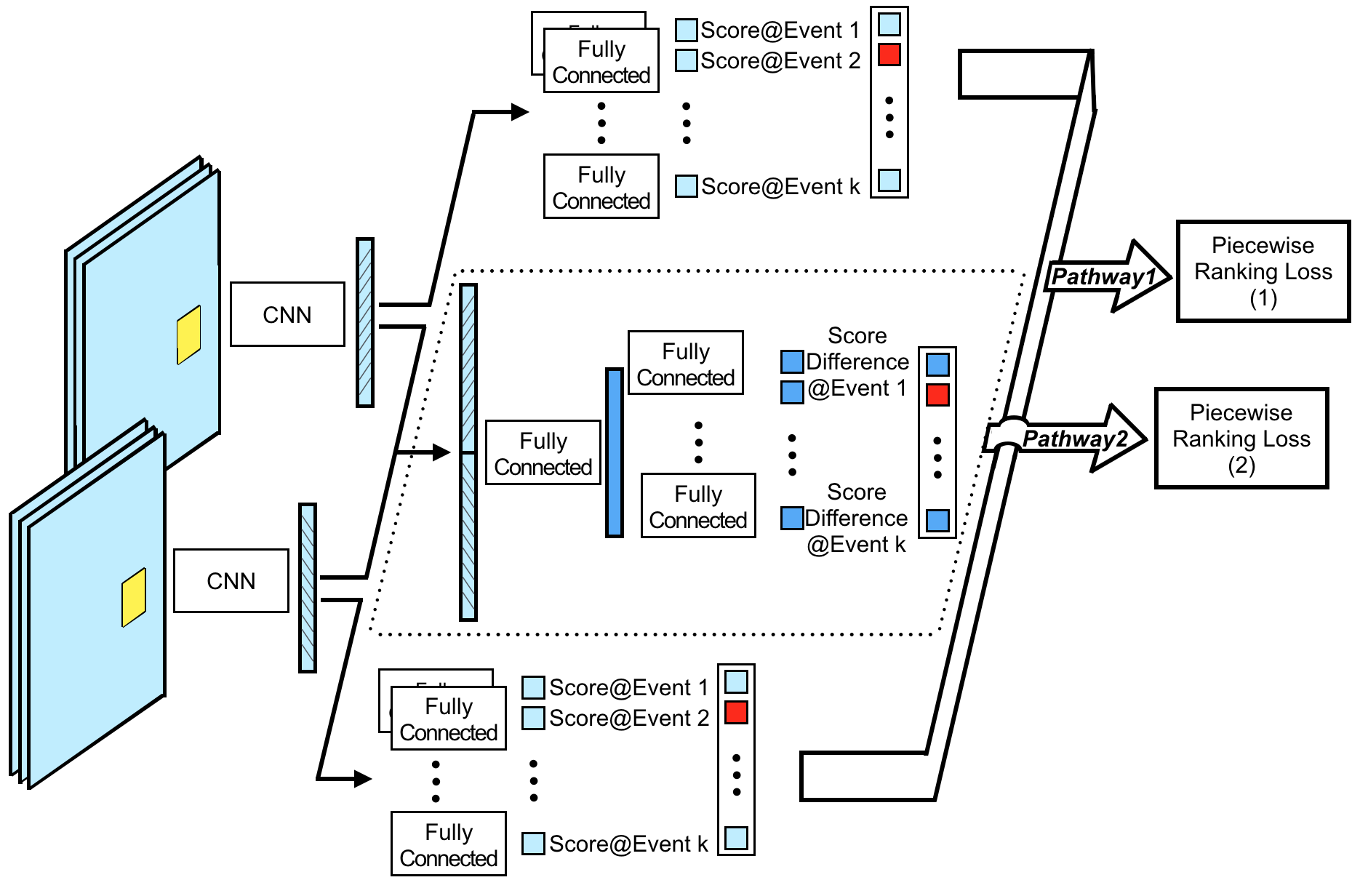}
\caption{Architecture of the event curation siamese network (Curation-Siamese) during training. } 
\label{siamese}
\vspace{-10pt}
\end{figure}

For event curation, we use a similar approach as in \cite{CVPR}, using Piecewise Ranking Loss to train a Siamese network to predict the importance score difference between an image pair given the ground-truth event type. The Siamese network outperforms a traditional CNN that directly predicts the absolute image importance score. Compared to the architecture in \cite{CVPR}, we added a pathway to directly predict the score difference between the image pair, rather than looking at the two images separately. This makes the training process faster and improves the results. The architecture is shown in Figure~\ref{siamese}, and the added pathway is \textit{\textbf{Pathway2}} in the middle. More details are presented in the supplementary material. 

For each training image pair, the ``ground-truth'' event is sampled from the label distribution, and used to gate the output and gradient of the network. We denote this network as \textbf{Curation-Siamese}.

\subsection{Event recognition networks}
One of the properties of  an ``event album'' that makes it distinct from a simple collection of images is that it is a sequence, and this provides us with the temporal relationship between the images. LSTMs have been successfully applied to sequential tasks \cite{showandtell, lstm, Graves, speech, lrcn}, and their ability to remember long-range temporal context is suitable for our task. Therefore, we use the LSTM network to capture the sequential information, in addition to a classical CNN that captures the visual features of a single image. 

We start with a CNN pre-trained on ImageNet \cite{caffe, imagenet}, and optionally fine-tune it on ML-CUFED to recognize the event type from a single image. We call this network \textbf{Recog-CNN}. We extract the high level CNN features for each image, and use them as the input features to train the LSTM network for album-wise event recognition. The LSTM network consists of a single LSTM layer as in \cite{lrcn}, a mean pooling layer, and a softmax layer. Cross entropy loss is used. We denote the LSTM as \textbf{Recog-LSTM}. More details about the structure of the LSTM network are presented in the supplementary material. The target for both the CNN and LSTM network is a one-hot encoding, but is sampled from the ground-truth label distribution.
\subsection{The iterative curation-recognition procedure}
\label{iterative-section}
For an ``event album'', more important images give us more information about the event type. For example, although a candle blowing image may only appear in an album once, it is a critical clue for revealing the event type of the album. However, as shown in \cite{CVPR}, the importance of an image is event-type dependent. Therefore, we propose an iterative update procedure to demonstrate that the image importance score and event recognition of an album can be used to improve each other's performance.

We denote an $N$-image album as $\mathbf{A}= \left \{I_{1}, ..., I_{N}  \right \}$. We assume $C$ different event types. The input to the algorithm is the output of the above three networks: 1) Recog-CNN produces an $N$-by-$C$ matrix $Q$, where each row is a probability distribution over event-types, given the image; 2) Recog-LSTM produces a 1-by-$C$ row vector of probabilities of event types, $\hat{p}$, given the image sequence; 3) Curation-Siamese produces a $N$-by-$C$ matrix $W$, where each row is the importance score of an image, given the event-type. The output of the algorithm is the $N$-dimensional column vector $v = [v_{1}, ..., v_{N}]^{T}$, which is the prediction of the importance score for all images in album $\mathbf{A}$, and the $C$-dimensional row vector $p = [p_{1}, p_{2},..., p_{C}]$, the distribution over the possible event types.


The iterative curation-recognition procedure is as follows:
\begin{enumerate}
\setlength\itemsep{-0.1em}
  \item \textbf{Re-weight Recog-CNN event prediction by image importance.}
  \begin{equation}
  {p'}(k+1) \propto (v^{T}(k))^{\alpha }\cdot  Q
  \label{eq1}
  \end{equation}
  where $v(k)$ is the $k$-th step prediction for all images' importance scores in album $\mathbf{A}$ (initialized to a uniform distribution) and $\alpha$ is a parameter that controls the strength of the importance score for the update. $p'$ denotes the updated  album event type prediction, normalized to a distribution. Thus, $p'$ is a distribution over event types that is the average of each image's event distribution weighted by the image's predicted importance score.

   \item \textbf{Combine event type predictions} with $\hat{p}$.
  \begin{equation}
  p(k+1) = \frac{1}{2}({p'}(k+1) + \hat{p})
  \label{eq2}
  \end{equation}
where $\hat{p}$ is the probability distribution of event types predicted by Recog-LSTM. Thus, $\hat{p}$ serves as an ``anchor'' for the prediction. 

  \item \textbf{Update image importance score with the updated event type distribution.}
\begin{flalign}
\begin{aligned}
& v(k+1)  
 \propto \left \{W \circ \textup{I}\left \{ p_{c}\geq m\cdot \max_{c'}(p_{c'}) \right \}_{(1, c)}\right \} \cdot  p(k+1)^T
\label{eq3}
\end{aligned}
\end{flalign}    
 where $W$ is the importance scores of all the images given the event type from Curation-Siamese,
 $\circ$ denotes element-wise multiplication of each row, and \textbf{I} is an indicator that returns 1 if its argument is true and 0 otherwise. Hence, \textbf{I} forms a binary mask that zeros out the importance scores for columns of $W$ that correspond to low-probability events, computed as a fraction $m$ (a parameter) of the maximum probability event. Thus, the updated image importance is the average of the importance score given different events, weighted by the event type probability. Elements of $v(k+1) $ are normalized to range from 0 to 1.
\end{enumerate}

By iterating Equations~\ref{eq1}-\ref{eq3}, we obtain the album-wise event prediction $p$ and image importance score prediction $v$. Note that this procedure is not guaranteed to converge, hence we set a maximum number of iterations, and if this maximum number is reached before convergence, the predictions for $p$ and $v$ are obtained by averaging over last three steps.

\vspace{-5pt}
\section{Experiments}
\vspace{-5pt}
In this section, we evaluate our approach for both event recognition and image importance prediction on ML-CUFED, and on another album-wise event recognition dataset Bossard \etal \cite{HMM} collected called PEC, we compare our event recognition result with Bossard \etal \cite{HMM} and Wu \etal \cite{pec22}.

\vspace{-5pt}
\subsection{Baselines}
\vspace{-5pt}
Our joint recognition-curation method produces two outputs: an album event type prediction, and an image importance prediction. 
For event recognition, we compare our result with the baseline from Recog-CNN. In addition, the intermediate result of our algorithm can also be compared with the final result to validate the necessity of each part of our system. Therefore, we compare our method with the following methods:
\vspace{-3pt}
\begin{itemize}
\setlength\itemsep{-0.17em}
  \item \textbf{CNN-recognition}: Use Recog-CNN to predict the event type for each image, and average the results. 
    \item \textbf{CNN-LSTM}: The prediction by Recog-LSTM. Note this uses Recog-CNN's feature representation as input.
  \item \textbf{CNN-Iterative}: Use the proposed method as described in Section~\ref{iterative-section}, but without step 2. Therefore, Recog-LSTM result is not involved. 
  \item \textbf{CNN-LSTM-Iterative}: Our full proposed method as described in Section~\ref{iterative-section}.
  \end{itemize}
\vspace{-3pt}

To evaluate our image importance prediction, we compare with several baselines:
\vspace{-5pt}
\begin{itemize}
\setlength\itemsep{-0.15em}
  \item \textbf{CNN-Noevent}: Train a Siamese Network to predict the importance score difference of an input image pair without any event-type information. All albums are considered to be part of the same ``uber'' event type.
  \item \textbf{CNN-Noevent(test)}: Use Curation-Siamese that is trained using the ground-truth event type information to gate the output error and back-propagation signal, while during testing, average the predicted importance score for all possible event types.  
    \item \textbf{CNN-LSTM-Iterative}: As above.
    \end{itemize}

\vspace{-8pt}
\subsection{Experimental details}
\vspace{-3pt}
\paragraph{Dataset} For ML-CUFED, we split the albums into training and test in a ratio of 4:1. The test set has 368 albums. To decide the hyper-parameters $(\alpha, m)$ in our iterative model, a validation set with 111 albums is extracted from the training set.  For the PEC Event Recognition Dataset \cite{HMM}, we use directly the test set consisting of 10 albums for each event type as described in \cite{HMM}, so that we can directly compare their results with ours.

\paragraph{Parameter setting} For both the Recog-CNN and the Curation-Siamese, we use two architectures: 8-layer AlexNet \cite{imagenet} and 101-layer ResNet \cite{resnet}. Both networks are pre-trained on ImageNet, and we fine-tune AlexNet on ML-CUFED. We use a similar training scheme to \cite{caffe}, but with a lower learning rate of 0.001. For Recog-LSTM, we use high-level features from the Recog-CNN as input. For fine-tuned AlexNet, we use \textit{fc7} layer features, while for ResNet, we use the \textit{pool5} layer features. We reduce the feature dimension to 512 with PCA. For the Recog-LSTM, the dimensionality of the LSTM is 512, and we use AdaDelta as the optimization method \cite{adadelta,theano1,theano2}. For Curation-Siamese, we follow the settings in \cite{CVPR} and choose the two margins as $m_{s} = 0.1$ and $m_{d}=0.3$. We set the number of iterations of our joint recognition curation algorithm to 10. Please find more analysis regarding the performance with respect to the iteration number in the supplementary material.

\paragraph{Evaluation} For event recognition on ML-CUFED, we use two metrics to evaluate the models: average accuracy and $\textup{F}_{1}$ Score.  $\textup{F}_{1}$ Score is the harmonic mean of precision and recall, and can account for multi-label ground-truth. Both accuracy and $\textup{F}_{1}$ are calculated with top-1 prediction. For event recognition on PEC, only average accuracy is used. For image importance prediction, we follow \cite{CVPR} using MAP@($t\%$) and Precision@($t\%$).  Precision is the ratio between the number of retrieved relevant images and the number of retrieved images. MAP is the averaged area under the precision-recall curve.

\subsection{Results on the ML-CUFED Dataset}
\subsubsection{Event-specific image importance}

For the image importance score prediction task, we compare our methods to several baselines in Table~\ref{aesthetic_table} using ResNet features. Results for AlexNet features are in the supplementary material. For an upper-bound of our method, we also show the result for CNN-GTEvent, where we assume the ground truth event type is known, and predict the importance score based on that.
CNN-GTEvent serves as the best result we can get when the event recognition stage is perfect. As shown in Table~\ref{aesthetic_table}, CNN-Noevent performs better than CNN-Noevent (test). This suggests the divergence of the importance prediction for different event types. CNN-LSTM-Iterative greatly outperforms the other two models, filling the gap between CNN-Noevent (test) and CNN-GTEvent by 79\%, and between CNN-Noevent and CNN-GTEvent by 62\% (averaged over 6 levels of $t$ and between MAP and P).


\begin{table}[t]
\centering
\scalebox{0.7}{
\begin{tabular}{c|cccccc|cccccc}
\thickhline
          & \multicolumn{6}{c|}{$\mathbf{MAP}@\mathbf{t\%}$}          & \multicolumn{6}{c}{$\mathbf{P}@\mathbf{t\%}$} \\ \thickhline
t\%       & 5     & 10    & 15    & 20    & 25    & 30    & 5  & 10  & 15 & 20 & 25 & 30 \\ \hline
Random & 0.113 & 0.161 & 0.211 & 0.256 & 0.303 & 0.350 & 0.044  &  0.090   &  0.142  &  0.193  & 0.243   & 0.298   \\
CNN-Noevent(test) & 0.272 & 0.330 & 0.380 & 0.434 & 0.483 & 0.530 & 0.167   &  0.256   & 0.327   & 0.379   &   0.432 & 0.476    \\
CNN-Noevent & 0.280 & 0.352 & 0.403 & 0.455 & 0.504 & 0.552  & 0.178   &  0.281   & 0.347   & 0.404   &   0.454 & 0.497 \\ 
CNN-LSTM-Iterative &\textbf{0.302}&\textbf{0.371}&\textbf{0.419}&\textbf{0.470}&\textbf{0.520}&\textbf{0.568}&\textbf{0.205}&\textbf{0.300}&\textbf{0.360}&\textbf{0.413}&\textbf{0.459}&\textbf{0.507} \\ \hline
\rowcolor{Gray}
CNN-GTEvent &0.309&0.383&0.432&0.482&0.529&0.573&0.205&0.311&0.373&0.428&0.472&0.512 \\ 
\thickhline
\end{tabular}
}
\vspace{5pt}
\caption{Comparison of event-specific image importance predictions with different methods using ResNet features. Evaluation metric here is $\text{MAP}@t\%$ and $P@t\%$. 
}
\vspace{-5pt}
\label{aesthetic_table}
\end{table}


\subsubsection{Event recognition}
Table~\ref{CUFED} shows the results of different methods for event recognition.  For an album with multiple labels, we deem it correctly predicted if the top-1 prediction is among the ground-truth event labels. As shown, ResNet features perform much better than AlexNet features. For both AlexNet and ResNet features, there is a performance gain over all three baselines.  
We can also observe that both iterative curation-recognition and LSTM method help to improve the final result. This suggests that both these types of information in an event album are helpful in deciding the event type of this album: image importance information, and album sequential information.

We compare our results with another CNN-based model in \cite{pec22}. Wu \etal \cite{pec22} use a fine-tuned AlexNet to extract image features, and aggregate the image features for album-wise prediction of event type. Here, we reimplement their approach for ML-CUFED, 
using both fine-tuned AlexNet features and ResNet features. Our model substantially outperforms theirs. 

\begin{table}[]
\centering
\scalebox{0.7}{
\begin{tabular}{c|cc|cc|cc}\thickhline
\textbf{Dataset}   & \multicolumn{4}{c|}{\textbf{ML-CUFED}}                                          & \multicolumn{2}{c}{\textbf{PEC}}                  \\ \thickhline
                   & \multicolumn{2}{c}{\textbf{Avg. Acc.}} & \multicolumn{2}{c|}{\textbf{F1-Score}} & \multicolumn{2}{c}{\textbf{Avg. Acc.}}            \\ \hline
Method             & AlexNet            & ResNet            & AlexNet            & ResNet            & AlexNet                  & ResNet                 \\ \hline
CNN-recognition    & 75\%               & 82.9\%            & 0.698              & 0.772             & 80.9\%                   & 84.5\%                 \\
CNN-LSTM           & 76.6\%             & 81.5\%            & 0.713              & 0.759             & 82.7\%                   & 85.5\%                 \\
CNN-Iterative      & 78\%               & 83.7\%            & 0.729              & 0.781             & 81.8\%                   & 86.4\%                 \\
CNN-LSTM-Iterative & \textbf{79.3\%}    & \textbf{84.5\%}   & \textbf{0.737}     & \textbf{0.786}    & \textbf{84.5\%}          & \textbf{87.9\%}        \\ \hline\hline
Wu \etal \cite{pec22} &   71.7\% & 83.4\%   &   0.662   & 0.773    &     *84.5\%      & *89.1\%        \\ \hline
SHMM \cite{HMM}               & \multicolumn{2}{c}{-}                  & \multicolumn{2}{c|}{-}                 & \multicolumn{2}{c}{*76.3\%} \\ \thickhline
\end{tabular}
}
\vspace{5pt}
\caption{Comparison of event-recognition models on ML-CUFED and PEC. Note that for the PEC result, our model is trained on ML-CUFED, while Wu \etal \cite{pec22}'s model and SHMM are trained on the PEC training set.}
\label{CUFED}
\vspace{-5pt}
\end{table}

\begin{figure}
\centering
\includegraphics[width=0.9\textwidth]{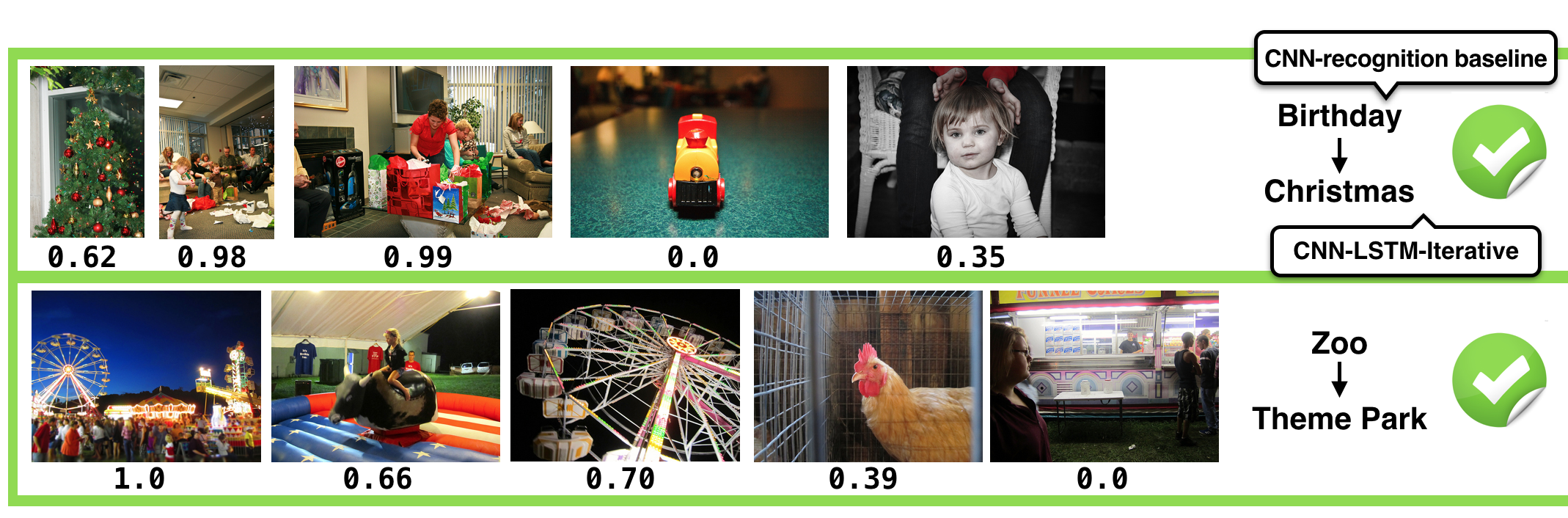}
\caption{Examples of our model's results on the ML-CUFED Dataset where the event type result is corrected by the iterative model, as shown
on the right. The images are in order of ground-truth importance, with the model's importance score below each image.
}
\label{resultcufed}
\end{figure}

In Figure~\ref{resultcufed}, we show our event curation and recognition result using ResNet with two examples in the ML-CUFED Dataset. The images are sorted in ground-truth importance order, and the predicted importance score is labeled below the image. To the right of the album, we show the event recognition of the album with the CNN-recognition method (before arrow),  and the CNN-LSTM-Iterative (after arrow). As shown, the event recognition is corrected with the CNN-LSTM-Iterative procedure. We show with more examples in the supplementary material.


\subsection{Results on the PEC Dataset}
\label{PEC_section}

\begin{figure*}[t]
\centering
\includegraphics[width=0.9\textwidth]{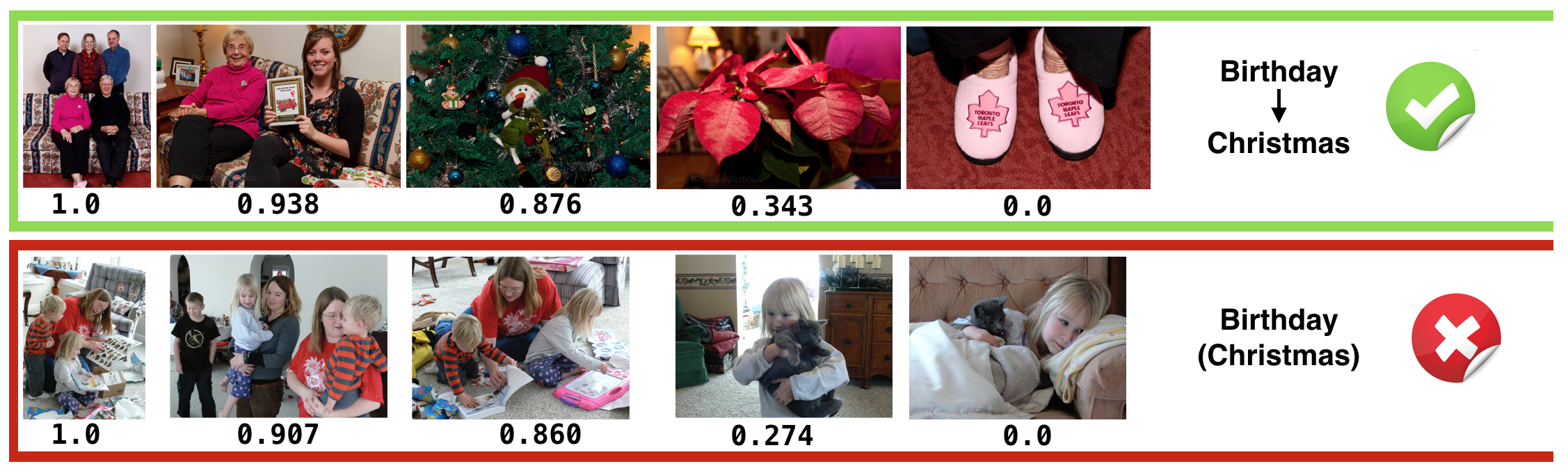}
\caption{Some results from the PEC dataset. For the first row, the prediction from CNN-recognition is wrong as shown in the label above, but is corrected in the final answer. For the second row, a case where the model fails to recognize a Christmas event is shown. Only a subset of images is shown for each album due to limited space. The images are sorted in predicted importance score order, and the predicted importance score of each image is shown below the corresponding image. }
\label{pec_result}
\end{figure*}

To show the generalizability of our algorithm, we compare our result with \cite{HMM} and \cite{pec22} on PEC.
The PEC dataset is an 807-album event dataset with 14 social event classes. There is no ground-truth importance score  in PEC, thus we cannot train our algorithm on it. Therefore, we use the model we trained on ML-CUFED and test the model on the PEC test set containing 10 albums each class.

PEC has several event types that are not contained in ML-CUFED, such as Saint Patrick's Day, Easter, and Skiing, and there are multiple event types that can map to single event type in ML-CUFED: Children's Birthday and Birthday can be mapped to single Birthday event in ML-CUFED. We provide the mapping from PEC label to ML-CUFED label in the supplementary material. Note that the mapping is not perfect,
and the noise in the mapping makes the performance of our method shown here a little poorer than it really is.

Due to the label changes, we recalculate the performance of Stopwatch HMM (SHMM) \cite{HMM} on the test data based on the confusion matrix they provided in the paper. For merged labels, the corresponding rows in the confusion matrix are merged. For missing labels, there are many possible approaches, and we follow the most loose one which assumes the best possible predictions:  Assume false positive on those labels will be correct predictions if those labels disappear. 

The comparison of different methods is shown in Table~\ref{CUFED}. Similar to the result on ML-CUFED, we observe consistent performance gain from both LSTM network and iterative updates. For the result of our reimplementation of \cite{pec22}, it is worth noticing that this model is trained on PEC, and it achieves current state-of-the-art result on PEC. Although our model is trained on ML-CUFED, it achieves very close performance with \cite{pec22}.



We show some examples of our recognition and event-specific image importance prediction result in Figure~\ref{pec_result}. There is no ground-truth labeling for the event-specific importance score, but we can look at the sample results qualitatively. From the first row, we can see that the model does not simply assign a high importance score to the characteristic Christmas tree which can distinguish the Christmas event , but predicts higher score to the family photo.
We show with more examples in the supplementary material.

\section{Conclusion}
In this work, we explore the problem of automatically recognizing and curating personal event albums. We attempt to solve the following two tasks jointly: recognizing the event type of an album, and finding the important images in this album. Specifically, the result from a CNN for image-wise event recognition, an LSTM Network for album-wise event recognition, and a Siamese Network for image importance prediction are integrated by a unified, iterated updating algorithm. We show that the joint algorithm significantly improves both image importance prediction and event recognition.

\bibliography{egbib}
\end{document}